\documentclass[10pt]{article}

\usepackage{arxiv}

\usepackage[utf8]{inputenc} 
\usepackage[T1]{fontenc}    
\usepackage{hyperref}       
\usepackage{url}            
\usepackage{booktabs}       
\usepackage{amsfonts}       
\usepackage{nicefrac}       
\usepackage{microtype}      
\usepackage{lipsum}
\usepackage{subcaption}
\usepackage{graphicx}

\usepackage{makecell}
\usepackage{multirow}
\usepackage{mathtools}
\usepackage{amsmath}
\usepackage{soulutf8}
\usepackage{color}
\usepackage{threeparttable}

\title{Challenges in Vessel Behavior and Anomaly Detection: From Classical Machine Learning to Deep Learning}

\author{
  Lucas May Petry\thanks{Corresponding author. Work developed during his visit to the Institute for Big Data Analytics at Dalhousie University.} \\
  Programa de Pós-Graduação em Ciências da Computação \\
  Universidade Federal de Santa Catarina (UFSC)\\
  Florianópolis, Brazil\\
  \texttt{lucas.petry@posgrad.ufsc.br} \\
  \And
  Amilcar Soares \\
  Institute for Big Data Analytics \\
  Dalhousie University \\
  Halifax, Canada \\
  \texttt{amilcar.soares@dal.ca} \\
  \And
  Vania Bogorny \\
  Programa de Pós-Graduação em Ciências da Computação \\
  Universidade Federal de Santa Catarina (UFSC)\\
  Florianópolis, Brazil\\
  \texttt{vania.bogorny@ufsc.br} \\
  \And
  Bruno Brandoli \\
  Institute for Big Data Analytics \\
  Dalhousie University \\
  Halifax, Canada \\
  \texttt{brunobrandoli@dal.ca} \\
  \And
  Stan Matwin \\
  Institute for Big Data Analytics \\
  Dalhousie University \\
  Halifax, Canada \\
  \texttt{stan@cs.dal.ca} \\
}

\begin{document}
\maketitle

\vspace{2em}
\begin{abstract}
The global expansion of maritime activities and the development of the Automatic Identification System (AIS) have driven the advances in maritime monitoring systems in the last decade.
Monitoring vessel behavior is fundamental to safeguard maritime operations, protecting other vessels sailing the ocean and the marine fauna and flora.
Given the enormous volume of vessel data continually being generated, real-time analysis of vessel behaviors is only possible because of decision support systems provided with event and anomaly detection methods.
However, current works on vessel event detection are ad-hoc methods able to handle only a single or a few predefined types of vessel behavior.
Most of the existing approaches do not learn from the data and require the definition of queries and rules for describing each behavior.
In this paper, we discuss challenges and opportunities in classical machine learning and deep learning for vessel event and anomaly detection.
We hope to motivate the research of novel methods and tools, since addressing these challenges is an essential step towards actual intelligent maritime monitoring systems.
\end{abstract}

\vspace{2em}
\textbf{NOTICE:} This is an extended version of the article \textbf{Challenges in Vessel Behavior and Anomaly Detection: From Classical Machine Learning to Deep Learning}, published by Springer in the proceedings of the 33rd Canadian Conference on Artificial Intelligence, available online: \url{https://www.researchgate.net/publication/340132071_Challenges_in_Vessel_Behavior_and_Anomaly_Detection_From_Classical_Machine_Learning_to_Deep_Learning}.
Please cite the original work.
\vspace{1em}

\section{Introduction}\label{sec:introduction}



\begin{figure}[h]
    \centering
    \includegraphics[width=\linewidth]{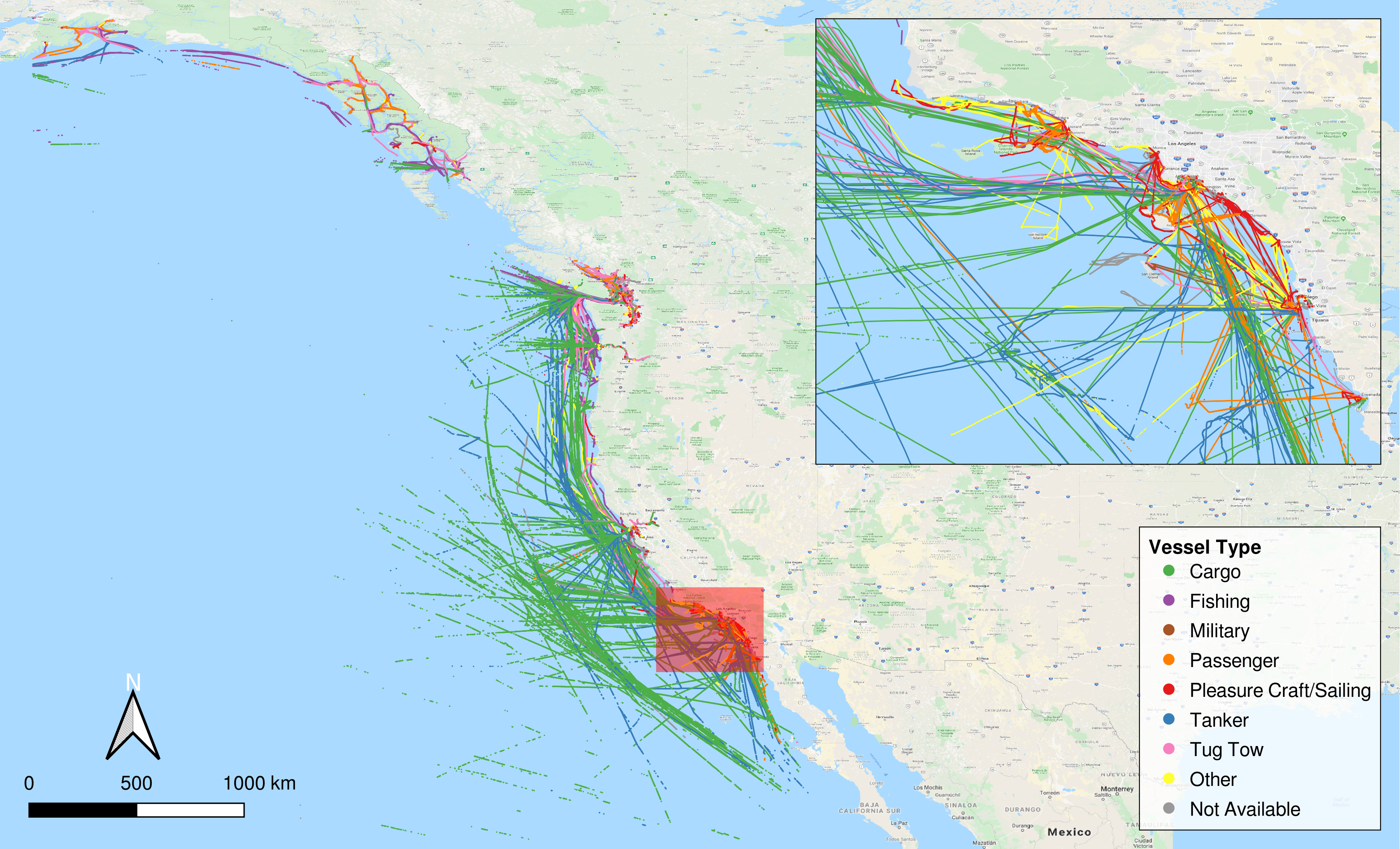}
    \caption{7,819,014 GPS points of the trajectories of 3,140 vessels in the US west coast during the third week of January, 2015, collected by the US Coast Guard.}
    \label{fig:ais_sample}
\end{figure}

Maritime transportation represents 90\% of all international trade volume \cite{vespe2012unsupervised} and more than 50,000 vessels sail the ocean every day \cite{Guillarme2013}.
The intercontinental trade of affordable food and manufactured goods is only possible thanks to the maritime traffic network.
Figure~{\ref{fig:ais_sample}} shows 7,819,014 GPS points of the trajectories of 3,140 vessels in the west coast of the United States (US) during only the third week of January 2015, collected by the US coast guard\footnote{Extracted from zones 6 to 11 at \url{https://marinecadastre.gov/ais/}.}.

The worldwide growth of maritime traffic and the development of the Automatic Identification System (AIS) has led to advances in monitoring systems for preventing vessel accidents and detecting illegal activities.
In addition, the integration of vessel traffic data with environmental and climatological data allows more complex analyses and a better understanding of the cause and effect of maritime events \cite{soares2019crisis}.
While preventing vessel accidents means saving money for shipping companies, from the environmental point of view, it also protects the marine fauna and flora from irreversible damage \cite{claramunt2017maritime}.

\begin{figure}
    \centering
    \includegraphics[width=.8\linewidth]{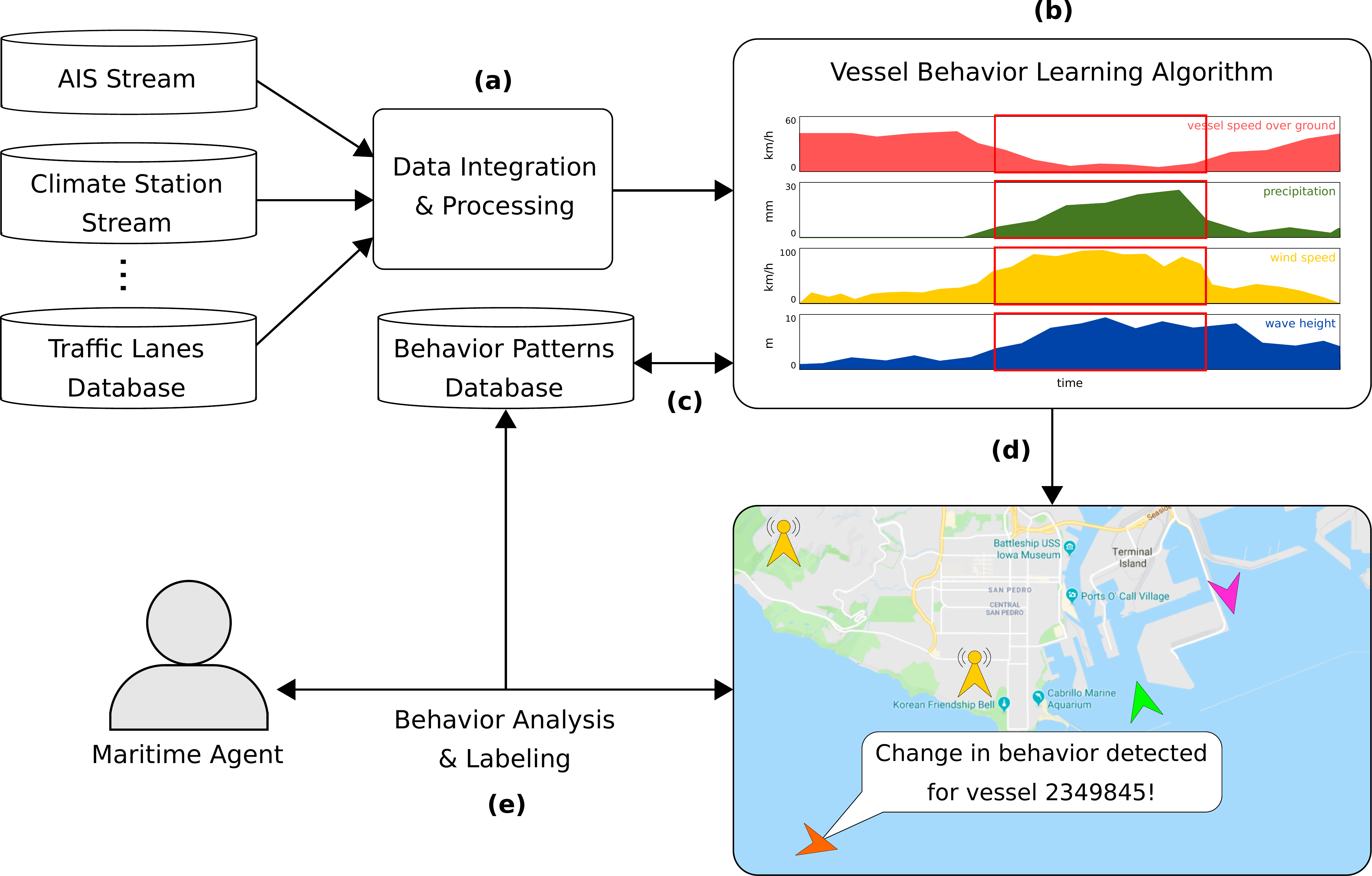}
    \caption{Overview of an ideal maritime monitoring system: (a) data constantly received from multiple sources are consumed, processed, and integrated; (b) a learning algorithm detects different vessel behaviors from these data, so that they can be (c) stored and (d) presented to the user (maritime agent); (e) the agent can analyze, understand, and label the detected behaviors, and also take any necessary actions.}
    \label{fig:overview_maritime_system}
\end{figure}

Real-time monitoring and analysis of vessel traffic can be overwhelming to maritime agents due to the high volume of data continuously generated.
Therefore, decision support systems are fundamental to enabling efficient and effective maritime control.
They can provide agents with interpretable semantic information about vessel behaviors as, for instance, assigning to a vessel the behavior of suspicious or illegal fishing activity for a particular time.
This information can direct the agents to focus on behaviors of interest, also known as events, further allowing quick responses to these events.
Figure~{\ref{fig:overview_maritime_system}} illustrates the overall architecture of the ideal maritime system we foresee in this paper.

The detection of events from AIS data has been the subject of study of several works in the literature \cite{lane2010maritime,van2012abstracting,patroumpas2015event,dividinosemantic,soares2019crisis,lei2019mining}.
In particular, some approaches have been proposed for detecting events such as changes in the speed \cite{patroumpas2017online,soares2019crisis,wen2019semantic} or in the course of vessels \cite{patroumpas2017online,wen2019semantic,varlamis2019network}, proximity of vessels to the coast or to other vessels \cite{lei2019mining}, drifting or loitering behavior \cite{pitsikalis2019composite}, illegal fishing \cite{patroumpas2015event} or possibly hazardous activity \cite{soares2019crisis}, among others.
However, to the best of our knowledge, most current works are ad-hoc approaches that do not learn from the data and are limited to detecting a restricted set of predefined events.
Such methods are not able to detect unforeseen events and also require the assistance of domain specialists for defining rules and thresholds that characterize each event.

Another aspect often ignored by previous research is the integration of data from different sources for analyzing vessel behavior.
Even though this can be advantageous to maritime systems, only a handful of works have addressed it \cite{kazemi2013open,terroso2016complex,soares2019crisis}.
For example, detecting small vessels heading towards ice-infested waters and that are not equipped for handling this situation allows the decision-maker to warn the captain in advance.
This behavior could be easily detected by matching vessel trajectories from AIS data with spatial regions of high ice concentration provided by climatological stations.
Such strategy avoids the deployment of a search and rescue mission, which might represent a high cost (e.g., lives, resources) to maritime authorities.
Nonetheless, the overwhelming amount of data continuously being generated by shipping companies, climate stations, and satellites, to name a few, poses challenges to maritime monitoring systems and their users.

In this work, we present research gaps and challenges in machine learning for detecting different types of vessel behavior, considering several constraints imposed by real-time data streams and the maritime monitoring domain.
We highlight the potential of exploiting machine learning techniques for maritime monitoring, as it has been shown to be fundamental for enabling cognitive smart cities, for instance, which is a scenario similar to ours with heterogeneous sensor data that requires real-time decision making systems~\cite{mohammadi2018enabling}.
Our contributions are as follows:
\begin{itemize}
    \item We introduce a 
    short survey of the state of the art for vessel event and anomaly detection;
    \item We present an extensive discussion of a few major topics with opportunities and challenges of research on vessel behavior detection with machine learning.
\end{itemize}

The rest of this paper is organized as follows.
In Section~\ref{sec:background}, we briefly introduce the terminology used in this work, as well as current advances in vessel behavior detection for maritime monitoring.
In Section~\ref{sec:challenges}, we present research challenges that remain unaddressed, indicating research directions for future works.
Finally, we present our final remarks in Section~\ref{sec:conclusion}.

\section{Background}\label{sec:background}

In this section, we describe the research advances that have been made for maritime monitoring concerning vessel behavior detection.
Before going into the details of related works, in the next section, we introduce the terminology used in this work.

\subsection{Terminology}\label{subsec:rw_terminology}

Existing works have often referred to the detection of vessel behaviors as vessel events or patterns.
In this work, we use the term \emph{behavior} to represent a set of rules or a data structure extracted from the vessel trajectory (AIS data) and other sensors (e.g., ground stations, buoys, satellites, etc.) in an environment.
An example of a behavior is a ship sailing in open seas, where its speed over ground is high (e.g., more than 15 knots), and its course over ground barely changes. 
An \emph{event}, is a behavior that is considered relevant to the application. 
An example of an event is a ship approaching a port area with high speed. 
Due to port regulations, ships are supposed to follow the speed limit posted under the port authority. 
Not complying with regulations may jeopardize the lives of people in the port area or can interfere with the local navigation.

In \cite{chandola2009anomaly}, anomaly detection is defined as \textit{``the problem of finding patterns in data that do not conform to expected behavior."}
In this work, however, we further extend this concept to facilitate the comprehension of existing approaches.
Given a method for anomaly detection, we define \textit{known anomalies} as the patterns that deviate from the normal behavior and that have a semantic meaning assigned by the method, i.e., the technique explicitly defines the properties and characteristics of these anomalies.
On the other hand, \textit{unknown anomalies} are defined as the patterns that deviate from the expected behavior but have no semantic meaning assigned by the detecting method.
For instance, the deviation of a vessel from a standard route may be an unknown anomaly.
From now on, we will refer to detected \textit{known anomalies} as identified vessel behaviors or events, which includes \textit{normal vessel behaviors}, and to \textit{unknown anomalies} simply as anomalies.

Unsupervised learning typically refers to the process of learning patterns from data instances that have not been previously labeled.
In this paper, we define a method as \textit{unsupervised} if it can learn from unlabeled data and does not require specific knowledge about the problem domain, neither for setup nor during run time.
Although it is difficult to draw sharp boundaries on the notion of human supervision, we consider a method as unsupervised only if domain-specific knowledge is not strictly necessary.
For instance, even though domain-specific knowledge may help one to define the number of clusters of a clustering algorithm, such parameter can be ``guessed" or even tuned without much effort.

\subsection{Detection of Vessel Events}
\label{subsec:rw_behavior_detection}

Several works have been proposed to address vessel behavior detection for maritime monitoring.
In \cite{terroso2016complex}, a method was proposed for the detection of abnormally low and high vessel speed, as well as possible collision situations.
In addition to AIS data, the method uses high-level descriptions of the weather conditions as a deciding factor for detecting abnormal low speed, i.e., it is expected that vessels reduce their speed under windy or rainy conditions, and such case does not configure abnormal low speed.
Other algorithmic approaches have been proposed for detecting Search And Rescue (SAR) patterns \cite{varlamis2018detecting}, conflicting vessel trajectories \cite{lei2019mining}, among others.

Fischer~\&~Bauer~\cite{Fischer2010} introduced an approach for the fusion of object observations produced by arbitrary heterogeneous sensors, called the Object-Oriented World Model (OOWM). 
Their goal is to fuse object observations that have been collected from multiple and heterogeneous sensors and platforms.
OOWM works, therefore, as an information source for application-level modules such as visualization, and higher-level methods like track analysis or behavior recognition.
Van~Hage~et~al.~\cite{van2012abstracting} proposed the Simple Event Model (SEM) ontology\footnote{\url{https://semanticweb.cs.vu.nl/2009/11/sem/}} for modeling generic events using Semantic Web standards, showing a use case for detecting vessel-related events in the maritime domain.
Simple events, such as moving and stopped, are detected algorithmically from raw data and modeled with SEM.
More complex events (e.g., vessel slowing down, anchored) are detected via inference in SEM with Prolog rules.

Soares~et~al.~\cite{soares2019crisis} presented a framework also based on Semantic Web technologies, namely Cross-queries over streams (CRISIS), for vessel event detection.
CRISIS integrates AIS data with data from different maritime and marine sensors for posterior reasoning with C-SPARQL.
The goal of CRISIS is to improve knowledge interoperability, and it is applied to the maritime ship traffic domain for discovering real-time traffic alerts by querying and reasoning across multiple streams.
Soares~et~al. define queries over the ontology to detect events, such as vessels over the speed limit, unequipped vessels heading towards ice-infested regions, vessels moving towards areas with extreme weather conditions, etc.

Another framework was designed by Kazemi~et~al.~\cite{kazemi2013open}, who considered the integration of AIS data with other available open data, such as weather conditions, port information, and social networks.
They modeled rules for 11 anomalous vessel behaviors in collaboration with maritime experts as, for instance, if a vessel destination is not in the port schedule if the vessel Estimated Time of Arrival (ETA) does not match the port ETA for the vessel, among others.
Although they present an overview of the framework architecture, not many details are provided as to how the data can be integrated and how the behaviors are detected.

Patroumpas~et~al.~\cite{patroumpas2015event} proposed a method for maritime event detection using the Event Calculus for Run-Time reasoning (RTEC), a logic programming language for reasoning about events.
Simple events such as a vessel stopped, turned, or changed its speed are recognized from a few trajectory points.
Long-lasting or more complex events are then recognized from these simple events via the definition of rules with RTEC.
The proposed complex events are a gap in communication, smooth turn, slow motion, illegal fishing activity, among others.
In \cite{patroumpas2017online}, the authors extended their previous work with improvements on the algorithms used to detect events, besides filtering noisy AIS data.


Lane~et~al.~\cite{lane2010maritime} presented a probabilistic approach to detect suspicious vessel behaviors solely based on AIS readings.
They model vessel deviation from standard route, AIS switch-off, unexpected port arrival, unusual proximity of vessels, and zone entry.
A Bayesian network is proposed for reasoning over the detected behaviors and further threat assessment.
Recently, Wen~et~al.~\cite{wen2019semantic} proposed the Semantic Model of Ship Behavior (SMSB), a semantic network for modeling vessel states and behaviors in a harbor scenario.
Besides modeling similar behaviors already described in previous works, SMSB also represents harbor-related behaviors such as entering or leaving an area, vessel anchored, approaching or departing from the harbor, joining main traffic flow, etc.
Similarly to \cite{lane2010maritime}, a dynamic Bayesian network is used to reason about vessel behavior.
Differently from previous works \cite{van2012abstracting,patroumpas2015event}, the Bayesian network can give credible results even when some information is missing.
However, a small sample of labeled data is required to learn the parameters of the network.

Other existing works tackle the problem of behavior detection in trajectory data, hence addressing more generic behavior patterns.
Siqueira~\&~Bogorny~\cite{de2011discovering} defined and presented an approach for detecting chasing behavior in trajectories, i.e., moving objects that are being chased by other objects.
Lettich~et~al.~\cite{lettich2016detecting} addresses the detection of avoidance behavior of a moving object concerning a region of interest.
In \cite{laube2005discovering}, the RElative MOtion (REMO) analysis concept is introduced, in which one trajectory is compared to all other trajectories in a dataset when looking for patterns.
REMO patterns are described using a formal notation similar to regular expression patterns.
REMO is analyzed with a case study of soccer players, where specific patterns regarding player performance and team coordination are found.

\subsection{Detection of Vessel Anomalies}

The use of adaptive Kernel Density Estimation (KDE) was proposed for vessel anomaly detection in \cite{ristic2008statistical}.
The normal behavior of vessels is modeled from historical training data based on the vessel's positions and speed measurements.
The method assumes the existence of a training set of usual vessel behavior, which is used as reference behavior when checking for anomalies on new incoming stream data.
Analogously, Kowalska~\&~Peel~\cite{kowalska2012maritime} proposed the use of Gaussian Processes (GPs) with historical AIS data for vessel anomaly detection without any supervision.
They analyze the heading and speed features of vessels separately and build a different GP model for every vessel type available.
As GPs are computationally expensive for large datasets, active learning is used to select a good sample of the whole dataset, which is then used to model normal vessel behavior. 
Anomalies are then detected based on the deviation in speed or heading of a vessel according to the expected normal behavior.
Riveiro~et~al.~\cite{riveiro2008improving} also proposed a methodology for anomaly detection with a Gaussian Mixture Model (GMM), but including interactive visualizations.
The presented approach relies on the user for validating anomalies detected by the GMM, so that the user can continuously refine the detection model.

Soleimani~et~al.~\cite{soleimani2015anomaly} presented an unsupervised method for ranking vessel trajectories based on their deviation from an optimal trajectory path, and so it can detect anomalous behavior.
This approach is based on the fact that vessels in open seas are usually interested in performing the shortest path to their destination, to reduce costs and/or meet deadlines in the case of cargo vessels, for instance.
A region of interest is first discretized into a grid of cells, and then the A* algorithm is used for finding the shortest trajectory path.
Finally, they extract geometrical features from the optimal and real trajectories (e.g., length, area under the curve, latitude, and longitude variation) for computing an anomaly score.

A few approaches in the literature have addressed the problem of vessel anomaly detection with neural networks.
Bomberger~et~al.~\cite{bomberger2006associative} proposed the training of a neural network to model the normal behavior of vessels and predict future vessel locations; hence, anomalies can be detected.
More recently, Nguyen~et~al.~\cite{nguyen2018multi} introduced a Recurrent Neural Network (RNN) model for vessel trajectory prediction and anomaly detection.
The vessel position, speed, and course over the ground are individually discretized and given as input features to the RNN model for further trajectory prediction, detection of vessel type, and abnormal vessel behavior.
Although neural network models have gained considerable popularity in the last decade, they often lack interpretability, which is an essential factor in maritime monitoring.

\subsection{Summary and Discussion}

Table~{\ref{tab:related_work}} outlines the main aspects of existing works for detecting vessel events and anomalies.
There is a clear distinction in the literature of works that perform event detection and general anomaly detection, though in many works these terms are used interchangeably.
For event detection, the majority of existing approaches is based on ad-hoc queries and rules for detecting different behaviors.
Although Lane~et~al.~\cite{lane2010maritime} and Wen~et~al.~\cite{wen2019semantic} use probabilistic models, the design of the model still relies on specific knowledge of the domain.
Hence, all approaches require some form of supervision, except the methods proposed in \cite{varlamis2018detecting,lei2019mining} that focus on a single type of behavior rather than a general method for behavior detection.

A good characteristic of current works on event detection is that they do not require pretraining on a dataset before the method can be used, which supports the fact that they do not actually learn from the data and rely heavily on human supervision.
In addition, only three works for event detection have explored the integration of AIS data with data from other sources (e.g. weather and ocean data), but to a limited extent.
Overall, the work of Soares~et~al.~\cite{soares2019crisis} is the most complete and easy to be extended including other vessel behaviors characterized by data from multiple sensors.

Works on anomaly detection, on the other hand, use mostly learning approaches for detecting anomalies that are not known beforehand.
Although the use of additional data sources can also be beneficial to anomaly detection, they have never been used in current approaches.
Despite the fact that most of these methods require no supervision, they generally lack in the extraction of knowledge and correlating the detected anomalies in a way that they can be interpreted by the user.

\begin{table*}
    \centering
    \caption{Comparison of the state of the art in vessel event and anomaly detection for maritime monitoring. We have omitted No's in Yes/No columns for greater legibility. While for event detection nearly all approaches are based on queries and rules defined by domain experts, most works for anomaly detection employ a learning algorithm to model normal vessel behavior, which is use as reference for detecting anomalies.}
    \label{tab:related_work}
    \resizebox{\textwidth}{!}{
    \begin{threeparttable}
    \begin{tabular}{clcccccc} \toprule
        & \multirow{1}{*}{\textbf{Work}} & \makecell{\textbf{Main approach}} & \makecell{\textbf{Unsupervised}} & \makecell{\textbf{No data}\\ \textbf{pretraining}} & \makecell{\textbf{Multi-sensor}\\ \textbf{data}} \\ \midrule
        \multirow{11}{*}{\makecell{\textbf{Event}\\ \textbf{detection}}} &
          Nilsson~et~al.~\cite{nilsson2008extracting}    & Queries/rules     &     & Yes &     \\
        & Lane~et~al.~\cite{lane2010maritime}         & Probability model &     & Yes &     \\
        & Van~Hage~et~al.~\cite{van2012abstracting}       & Queries/rules     &     & Yes &     \\
        & Kazemi~et~al.~\cite{kazemi2013open}           & Queries/rules     &     & Yes & Yes \\
        & Patroumpas~et~al.~\cite{patroumpas2015event}      & Queries/rules     &     & Yes &     \\
        & Terroso~et~al.~\cite{terroso2016complex}       & Algorithm/rules   &     & Yes & Yes* \\
        & Patroumpas~et~al.~\cite{patroumpas2017online}     & Queries/rules     &     & Yes &     \\
        & Varlamis~et~al.~\cite{varlamis2018detecting}**  & Algorithm/rules   & Yes & Yes &     \\
        & Wen~et~al.~\cite{wen2019semantic}          & Probability model &     & Yes &     \\
        & Lei~\cite{lei2019mining}**          & Clustering        & Yes & Yes &     \\
        & Soares~et~al.~\cite{soares2019crisis}         & Queries/rules     &     & Yes & Yes \\ \midrule
        \multirow{9}{*}{\makecell{\textbf{Anomaly}\\ \textbf{detection}}} &
          Bomberger~et~al.~\cite{bomberger2006associative} & Neural network & Yes &     &     \\
        & Ristic~et~al.~\cite{ristic2008statistical}    & KDE & Yes &     &     \\
        & Riveiro~et~al.~\cite{riveiro2008improving}     & Gaussian model &     &     &     \\
        & Kowalska~\&~Peel~\cite{kowalska2012maritime}     & Gaussian model &     &     &     \\
        & Pallotta~et~al.~\cite{pallotta2013vessel}       & Clustering + Probability model & Yes & Yes &     \\
        & Soleimani~et~al.~\cite{soleimani2015anomaly}     & Algorithm/rules & Yes & Yes &     \\
        & Nguyen~et~al.~\cite{nguyen2018multi}          & Neural network & Yes &     &     \\
        & Varlamis~et~al.~\cite{varlamis2019network}      & Clustering & Yes &     &     \\
        \bottomrule
    \end{tabular}
    \begin{tablenotes}
        \item[*] They only use the weather description (e.g. sunny, rainy) as a deciding factor about detected abnormal low speed behavior.
        \item[**] Only a single type of behavior/event is detected.
    \end{tablenotes}
    \end{threeparttable}}
\end{table*}

\section{Research Challenges and Opportunities}\label{sec:challenges}

In this section, we explore the challenges and research gaps that are currently faced by vessel behavior monitoring techniques.
We address three main tasks related to vessel behavior detection:
\begin{enumerate}
    \item The actual detection of different vessel behaviors or behavior changes;
    \item Identifying and relating recurrent behaviors, i.e., determine if and when a behavior has been previously detected;
    \item Providing the user with means for interpreting and analyzing the detected vessel behaviors.
\end{enumerate}

Although a large volume of raw data is available (i.e., data extracted directly from sensors such as AIS, climatological stations, that have not been processed), it presents some limitations.
Particularly, we discuss about three main issues associated to the data that can be challenging for machine learning methods:
\begin{enumerate}
    \item The data has limited or no labels about vessel behaviors. Labels represent the expected output of a predictor and enable the use of supervised learning algorithms. Their use in vessel behavior detection, however, is currently limited due to the lack of labels;
    \item Often, there is no knowledge whatsoever of what are the behaviors or labels present in the data;
    \item Finally, integrating streaming data from multiple data sources may result in highly-dimensional data points, which pose new challenges to behavior detection methods.
\end{enumerate}

In the following sections, we provide more details regarding the aforementioned tasks and data issues, which have not been addressed or only partially addressed by state-of-the-art approaches.
We also connect the gaps present in vessel tracking for maritime surveillance with other areas of research that can potentially contribute to the advancement of vessel monitoring with machine learning.


\subsection{Behavior Detection: Supervised vs. Unsupervised}
\label{subsec:supervised_vs_unsupervised}

Most of the existing works for behavior detection in vessel traffic monitoring do not learn from the data.
Instead, algorithms are proposed to detect a restricted set of vessel activities via rules and thresholds defined beforehand.
Defining such rules requires the knowledge from domain specialists, which may introduce considerable human bias in the analysis.
Moreover, defining strict attribute thresholds (e.g., vessel speed above a specified threshold configures high-speed) can be difficult and may induce the identification of irrelevant events.

Although the sole use of supervised learning techniques may seem appealing, they can also limit the detection of events to a predefined set, i.e., the classification labels.
Besides the difficulty or even impossibility of enumerating all possible different vessel behaviors, system and user requirements continuously change.
For instance, while a maritime agent may be mostly interested in detecting illegal fishing activity, another user might be concerned about vessels speeding or with conflicting trajectories near the coast.
With that in mind, we believe that research on unsupervised learning techniques is promising for taking vessel event detection research to the next level.

A closely related research topic that has already been explored in the literature is the detection of concept drift or change in time series data \cite{bifet2007learning,qahtan2015pca,sethi2016grid}.
In streaming data, concept drift is commonly referred to as the detection of significant changes in the data distribution \cite{gonccalves2014comparative}.
Several works for change detection in data streams were designed to work along with a supervised learning model, detecting drifts based on the error rate of the learner \cite{bifet2007learning,bach2008paired,ross2012exponentially,sethi2016grid}.

Unsupervised approaches have also been proposed for detecting concept drift \cite{bifet2007learning,lee2012detection,kuncheva2013pca,qahtan2015pca}.
However, they also exhibit some drawbacks, such as being limited to univariate data \cite{bifet2007learning}; simply detecting changes based on individual feature correlation \cite{lee2012detection} instead of analyzing dependency relationships between features and how they characterize different behaviors; and even lacking direct interpretability of the detected changes~\cite{kuncheva2013pca,qahtan2015pca}.
In spite of their limitations, these methods can provide a solid starting point for future works on unsupervised behavior change detection and posterior event notification.


Another limitation of many existing works is that they have mainly analyzed the individual behavior of vessels, but some anomalies may only be detected when analyzing a group of vessels.
In fact, similarly to the discussion presented in \cite{miller2019towards}, understanding collective movement patterns of vessels can be an easier task than analyzing individual trajectory behaviors, and it may also assist in the assessment of individual behaviors.
For instance, a vessel switching off the AIS device on purpose to hide some illegal activity can be detected by analyzing the AIS messages captured in an area where several others were transmitting at that time.
Another example is given in Figure~\ref{fig:ais_sample}, where it can be noticed that mainly Cargo and Tanker vessels sail in open seas, while Passenger and Pleasure Craft/Sailing vessels move for the most part along the coast.

Clustering techniques have already been used for analyzing ship behavior~\cite{zhou2019,lei2019mining}, but to a limited extent and for very specific purposes.
Analogous to well-known clustering-based trajectory mining algorithms~\cite{lee2007trajectory,lee2008traclass}, clustering patterns could be used as features for characterizing individual vessels, and thus support behavior detection.
However, the streaming scenario and real-time constraints present in maritime monitoring must be observed.

\subsection{Identifying Recurrent Behavior Patterns}
\label{subsec:recurrent_behavior}

\begin{figure*}[b]
    \centering
    \includegraphics[width=1\linewidth]{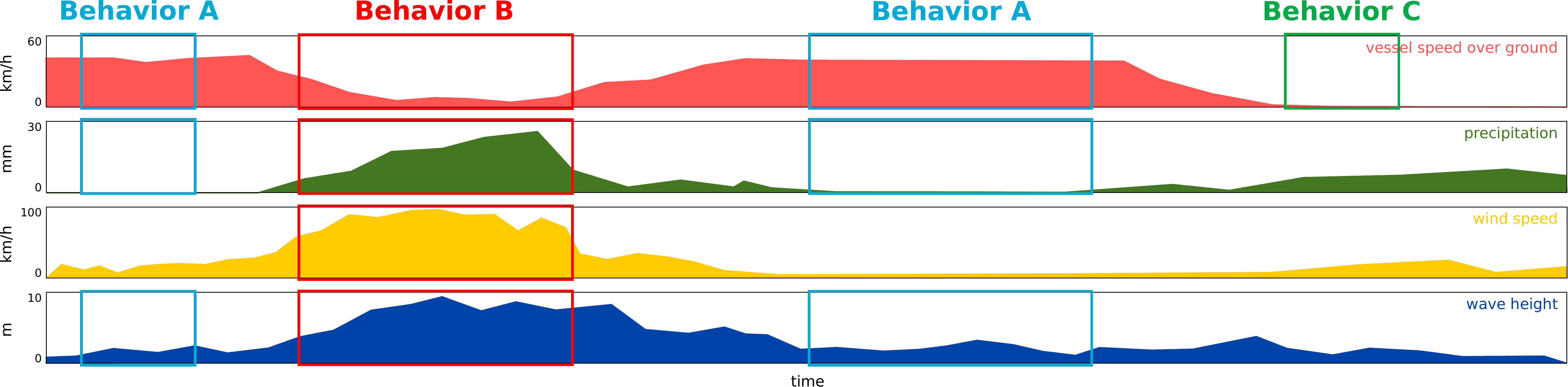}
    \caption{Toy streaming data of a maritime scenario with AIS (vessel speed over ground), environmental (precipitation and wind speed), and ocean data (wave height). Highlighted behaviors show regular vessel movement (A), vessel going through a tropical storm (B), and vessel anchored (C). Different behaviors may be described by different features.}
    \label{fig:behavior_sensor_example}
\end{figure*}

Detecting multiple instances of the same behavior is an essential factor in maritime monitoring.
Besides avoiding multiple analyses of the same behavior by agents, it allows agents to have a higher-level picture of behavior patterns of a single vessel or even of a group of vessels in a region.
Figure~\ref{fig:behavior_sensor_example} illustrates the detection of different vessel behaviors based on AIS, climate, and environmental data.
We highlight the detection of the same behavior (A) in two different points of the series.

In previous works for vessel event detection, identifying multiple occurrences of the same behavior was a trivial task, since the detection process consisted mostly of a query \cite{soares2019crisis} or an algorithm for detecting a single behavior \cite{terroso2016complex,lei2019mining}.
On the other hand, in the unsupervised setting, a pattern or behavior needs to be characterized in a way that multiple occurrences of the same pattern can be identified.
This can be a challenging task given the high number of features that may exist and depending on the length of the pattern.
Although concept drift or behavior change detection methods can indicate boundaries for different types of behavior in streaming data, they do not provide a way for detecting recurrent behavior, i.e., indicating how different time series segments relate to each other.

To the best of our knowledge, the only unsupervised learning method proposed for detecting different and recurrent behaviors in time series data is the Toeplitz Inverse Covariance-based Clustering (TICC), introduced by Hallac~et~al.~\cite{hallac2017toeplitz}.
TICC segments multivariate sensor data into sequences of states or clusters (i.e., behavior patterns), representing each behavior as a Markov Random Field (MRF).
However, the number of clusters is fixed and must be defined by the user, meaning that the number of different behaviors present in the data should be known a priori.
Additionally, TICC is not directly suitable for streaming data, as it assumes that all data is available at the same time, and it requires a few iterations of the algorithm for convergence.
Although TICC has some drawbacks, we believe that future research could take advantage of the method for recurrent behavior detection as, for instance, use MRFs to represent and identify the same behavior in different trajectory segments.


\subsection{Towards Interpretable Behavior Patterns}
\label{subsec:feature_selection}

Doshi~\&~Kim~\cite{doshi2017towards} define interpretability in the context of machine learning as ``the ability to explain or to present in understandable terms to a human."
In order to enable maritime agents to make data-driven decisions about suspicious or dangerous vessel activity, discovered behavior patterns must be interpretable, i.e., understandable to the agents.
We elucidate two main aspects of interpretability, one related to the interpretation of the model and the other regarding the characterization of the detected behavior patterns.
The former provides users with the reasoning behind the decisions taken by the model, i.e., why a trajectory segment and other pieces of data characterize a specific behavior.
On the other hand, the latter provides users with human-readable information of what specifically distinguishes that behavior from the rest of the data, which is sometimes also encoded in the model (e.g., decision trees).
Moreover, interpretability may assist in the detection of recurrent behaviors if they can be explicitly characterized based on feature observations.

Interpretability is, perhaps, the main advantage of a few existing works, since rules explicitly define events related to vessel behavior.
In contrast to these approaches, detecting interpretable patterns can be challenging for unsupervised machine learning algorithms.
For vessel behavior detection, we conjecture that the interpretability of the model may be negligible if guarantees are given about the characteristics of the detected behaviors.
For instance, if a behavior can be represented as an MRF, and it is always defined by the same variable dependency graph, the user might not be interested in the details of the underlying model as long as this structure is guaranteed for all future occurrences of the same behavior.
Thus, detection methods that are based on abstractions of the real observed features (e.g., \cite{qahtan2015pca,manzoor2018xtream}) can be a feasible option, which we discuss in Section~\ref{subsec:sparsity_dimensionality}.
Afterwards, other techniques could be exploited for correlating and providing interpretability of different behaviors from the real observed variables.

\subsection{Diving Into Deep Learning}
\label{subsec:deep_learning}

Deep learning models have often been set aside in favor of linear models, because of the claimed lack of interpretability that they have~\cite{lou2012intelligible}.
We believe that for a similar reason and added the lack of labeled data, only very few works have addressed anomaly detection in the maritime domain with deep learning~\cite{protopapadakis2017stacked,nguyen2018multi,zhao2019maritime}.
To the best of our knowledge, no work has addressed the detection of specific vessel behaviors with deep learning techniques.
More recently, however, a few works have been proposed to assist in the visual interpretation of these models \cite{simonyan2013deep,zhou2016learning,zhang2018interpretable}, while others have even questioned previous claims over the interpretability of deep models \cite{lipton2016mythos}.


Computer vision is, undeniably, the field that has experienced the most advances with deep learning, and Convolutional Neural Networks (CNNs) are nowadays widely used for image classification~\cite{krizhevsky2012imagenet}.
An intuitive approach one could think of is the use of satellite imagery data with CNNs for detecting ship anomalies.
However, it may be difficult or even impossible to detect certain vessel behaviors from satellite imagery, which limits the use of modern deep learning methods exclusively based on image data~\cite{nyman2019techno}.
Also, obtaining satellite images is generally more difficult and expensive in comparison with AIS data.
In this context, other authors have also explored the use of images obtained from Unmanned Aerial Vehicles (UAVs).
For instance, Xiu~et~al.~\cite{xiu2019}~conducted a study with UAVs for vessel identification by using AIS data fused with high-resolution aerial images.
On the other hand, CNNs have already been used for trajectory classification \cite{dabiri2018inferring} and prediction \cite{lv2018t}, based on trajectory features such as speed, acceleration, and bearing rate, which is also extensively available in AIS data.
Therefore, CNNs, together with visual techniques for interpretability \cite{simonyan2013deep,zhang2018interpretable} could be further exploited for vessel behavior and anomaly detection.


\subsection{Addressing Data Sparsity and High-Dimensionality}
\label{subsec:sparsity_dimensionality}

In Figure~\ref{fig:behavior_sensor_example}, we also exemplify three different events detected from streaming AIS, climate, and ocean data.
Notice that while regular vessel movement (A) is identified from the vessel speed over ground, precipitation, and wave height features, the vessel anchored (C) can be determined solely by the vessel speed over ground.
Different subsets of features may determine various types of behaviors, and this, added to the high number of features that might be available, presents another level of complexity to unsupervised learning algorithms.

Some works have addressed the high-dimensionality issue for change or anomaly detection in multidimensional streaming data~\cite{qahtan2015pca,manzoor2018xtream}.
They have done so by mapping the feature space to a low-dimensional abstract feature space with Principal Component Analysis (PCA), from which changes are detected.
Another path for dimensionality reduction lies in the use of autoencoders or variational autoencoders~\cite{kingma2013auto}.
While PCA is essentially a linear transformation, autoencoders are able to model complex non-linear relationships that may be present in the data.
In addition, autoencoders can be easily combined with Recurrent Neural Networks (RNNs) for sequence-aware dimensionality reduction.
However, caution should be exercised in the use of autoencoders since they are computationally more expensive than PCA and prone to overfitting.

Vessel behavior detection can also benefit from advances in Natural Language Processing (NLP), especially in dimensionality reduction.
For instance, a popular technique used for reducing the dimensionality and extracting meaningful similarity-based representations of textual data was proposed by Mikolov~et~al.~\cite{mikolov2013efficient}, also known as \textit{word2vec}.
The potential applicability of word2vec in other areas has been shown in multiple works, including in the field of trajectory data mining~\cite{gao2017identifying,esuli2018traj2user}.
Future works could take advantage of this and other NLP approaches for embedding streaming data from AIS and other sensors, or even for automatically interpreting and integrating port regulations with AIS data for vessel behavior analysis.
This would allow the monitoring system to adapt to new regulations without requiring human supervision for deploying new behavior rules.

\subsection{Big, Yet Limited Data}
\label{subsec:big_limited_data}

The growth of maritime activity led to advances in AIS technology, and such developments resulted in large volumes of data being generated every day.
As we mentioned in the beginning of Section~\ref{sec:challenges}, although a large amount of raw data is available, it lacks labels.
Labels are a piece of valuable information for researchers and machine learning algorithms. 
Labels represent the expected output of a predictor, hence supervised learning algorithms can learn a function that maps inputs to outputs based on provided examples. 
For event detection, labels could represent the different types of vessel behavior that should be detected by an algorithm.
Labeled data not only help algorithms to learn better how the observed variables explain different patterns of behaviors but also ease the validation and evaluation of algorithms during research.
However, labeling data is difficult since it requires knowledge from domain experts, which is time-consuming and, therefore, expensive.

The lack of labels has guided the focus of current research either to detecting a single type of vessel behavior in an unsupervised manner \cite{varlamis2018detecting,lei2019mining}, or to proposing the detection of different events via predefined behavior rules, materialized in the form of data queries~\cite{patroumpas2015event,patroumpas2017online,soares2019crisis}.
Future works could take advantage of the knowledge described in previous approaches for labeling data, to provide input-output examples to machine learning algorithms, as well as ground truth for evaluating novel approaches.
However, this should be done carefully as a strong human bias can be passed on to algorithms because much of the knowledge and rules presented in current works are a product of human judgment. 

A few tools were proposed in the literature to assist the user in the process of labeling movement data such as TagMyDay~\cite{TagMyDay13} and VISTA~\cite{soares2019vista}.
TagMyDay provides a web interface to upload personal trajectories and allows the user to annotate each segment with the activity performed by the generating user.
VISTA was designed to assist the user in the trajectory annotation process in a multi-role user environment.
A session manager creates a tagging session selecting the trajectory data and the semantic contextual information.
However, none of these works proposed a strategy to learn from the labels provided by the user interactions with the system.
After providing a few labels, a model could learn the behavior patterns describing each segment labeled by the user and suggest new occurrences of it in unseen trajectory segments.
The user could accept or refine the suggestions to improve the model so that with time, fewer interactions between the user and the system are needed.

Another possible research path to alleviate the label shortage problem, which remains unexplored in the maritime domain, is the use of transfer learning for boosting the learning process from few labeled examples, also known as few-shot classification \cite{vinyals2016matching,snell2017prototypical}.
Transfer learning approaches have been successfully applied with deep learning in the image classification domain~\cite{chen2019a}, where knowledge from another related domain is used as a means for enabling learning from a set of few examples.
On the other hand, instead of focusing on supervised techniques for learning from limited data, research could concentrate on synthesizing new behavior data with Generative Adversarial Networks~\cite{goodfellow2014generative}, for example, for enhancing the performance of other supervised models.

\section{Summary and Final Remarks}\label{sec:conclusion}

Even though maritime monitoring has experienced significant progress in the last decade, most of existing works do not take full advantage of machine learning techniques for vessel behavior detection.
In fact, existing research on behavior detection has focused on the proposal of queries and rules, as well as ad-hoc algorithms, for detecting specific types of behavior.
We argue that this methodology inhibits further advances to more general behavior detection approaches, constraining monitoring systems to function only under frequent human supervision.

\begin{figure}[h]
    \centering
    \includegraphics[width=.8\linewidth]{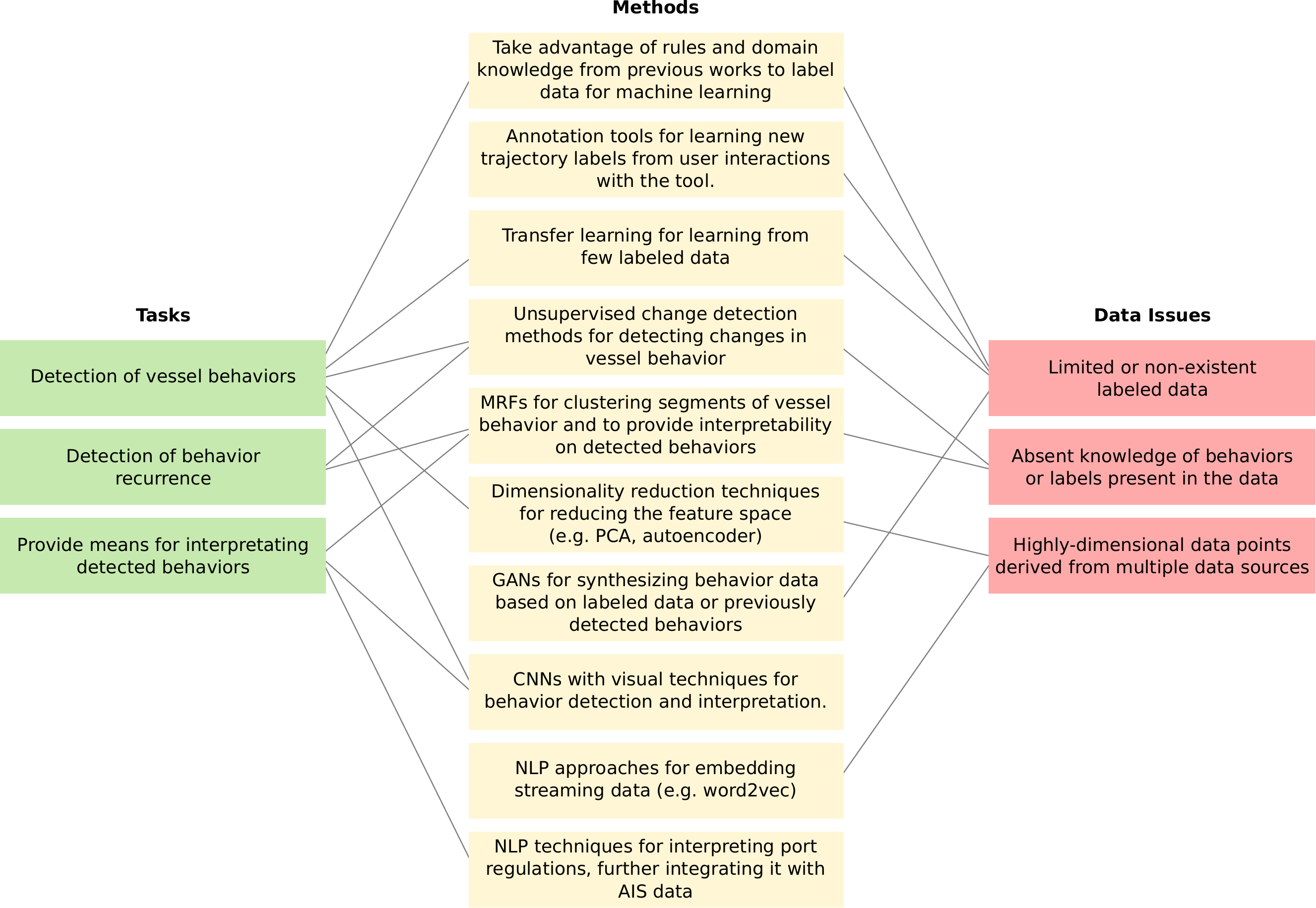}
    \caption{Summary of challenges according to existing tasks and data issues, as well as potential methods for addressing each of these challenges.}
    \label{fig:challenges_summary}
\end{figure}

In this paper, we presented the main research gaps in the field, indicating points of improvement and opportunities for future works.
In Figure~\ref{fig:challenges_summary}, we present a summary of the discussed opportunities of research according to the existing tasks and data issues for vessel behavior detection via machine learning.
We hope to instigate the development of new algorithms, methods, and tools for ship behavior monitoring, as many aspects of it still remain unaddressed.
Although there are some unique aspects of vessel behavior, other areas of research in trajectory pattern mining can potentially take advantage of improvements made in the field.
Finally, we believe that this work is another step towards actual intelligent maritime monitoring systems.

\section*{Acknowledgments}

This study was financed in part by Coordenação de Aperfeiçoamento de Pessoal de Nível Superior - Brasil (CAPES) - Finance Code 001 and through the research project Big Data Analytics: Lançando Luz dos Genes ao Cosmos (CAPES/PRINT process number 88887.310782/2018-00).
This work was also supported by Conselho Nacional de Desenvolvimento Científico e Tecnológico (CNPq), Fundação de Amparo a Pesquisa e Inovação do Estado de Santa Catarina (FAPESC) - Project Match (co-financing of H2020 Projects - Grant 2018TR 1266), and the European Union’s Horizon 2020 research and innovation programme under Grant Agreement 777695 (MASTER).
The authors also acknowledge the support of the Natural Sciences and Engineering Research Council of Canada (NSERC) and of Global Affairs Canada (GAC) - Emerging Leaders in the Americas Program (ELAP) for this research.

\bibliographystyle{unsrt}
\bibliography{references}

\end{document}